\pdfoutput=1

\documentclass[11pt]{article}

\usepackage{acl}

\usepackage{times}
\usepackage{latexsym}

\usepackage[T1]{fontenc}

\usepackage[utf8]{inputenc}

\usepackage{microtype}

\usepackage{times}
\usepackage{latexsym}
\usepackage{booktabs}
\usepackage{tabularx}
\usepackage{todonotes}
\usepackage{multirow}
\usepackage{url}
\usepackage{amsmath}
\usepackage{mathtools}
\usepackage{amssymb}
\usepackage{amsthm}
\usepackage{tablefootnote}
\usepackage{caption}
\usepackage{subcaption}
\usepackage{bbm}
\usepackage{comment}
\usepackage{xspace}
\usepackage{algorithm}
\usepackage{proof}
\usepackage{stmaryrd}
\usepackage{bm}
\usepackage{pifont}

\newcommand{\answerTODO}[1][]{\textcolor{red}{\bf [TODO]}}

\newcommand{\rngqa}{\textsc{RnG-KBQA}}
\newcommand{\grail}{\textsc{GrailQA}}

\newcommand{\webqsp}{\textsc{WebQSP}}

\title{\rngqa{}: Generation Augmented Iterative Ranking for \\ Knowledge Base Question Answering}

\author{
 Xi Ye$^\diamondsuit$\thanks{\,\, Work done during internship at Salesforce Research.} \quad  Semih Yavuz$^\spadesuit$ \quad  Kazuma Hashimoto$^\spadesuit$ \quad Yingbo Zhou$^\spadesuit$ \quad Caiming Xiong$^\spadesuit$ \\
 $^\diamondsuit$ Department of Computer Science, The University of Texas at Austin \\
 $^\spadesuit$ Salesforce Research \\
  $^\diamondsuit${\texttt{xiye@cs.utexas.edu} } \\
  $^\spadesuit${\texttt {\{syavuz,k.hashimoto,yingbo.zhou,cxiong\}@salesforce.com}}
}

\begin{document}
    \maketitle
    \begin{abstract}

    
    Existing KBQA approaches, despite achieving strong performance on i.i.d. test data, often struggle in generalizing to questions involving unseen KB schema items. Prior ranking-based approaches have shown some success in generalization, but suffer from the coverage issue. We present RnG-KBQA, a \textbf{R}ank-a\textbf{n}d-\textbf{G}enerate approach for KBQA, which remedies the coverage issue with a generation model while preserving a strong generalization capability. Our approach first uses a contrastive ranker to rank a set of candidate logical forms obtained by searching over the knowledge graph. It then introduces a tailored generation model conditioned on the question and the top-ranked candidates to compose the final logical form. We achieve new state-of-the-art results on \grail{} and \webqsp{} datasets. 
    In particular, our method surpasses the prior state-of-the-art by a large margin on the \grail{} leaderboard. In addition, RnG-KBQA outperforms all prior approaches on the popular \webqsp{} benchmark, even including the ones that use the oracle entity linking. 
    The experimental results demonstrate the effectiveness of the interplay between ranking and generation, which leads to the superior performance of our proposed approach across all settings with especially strong improvements in zero-shot generalization.\footnote{Code available at \href{https://github.com/salesforce/rng-kbqa}{https://github.com/salesforce/rng-kbqa}.}
    
    
\end{abstract}
    \section{Introduction}

\begin{figure}[t]
    \centering
    \includegraphics[width=\linewidth,trim=370 420 370 420,clip]{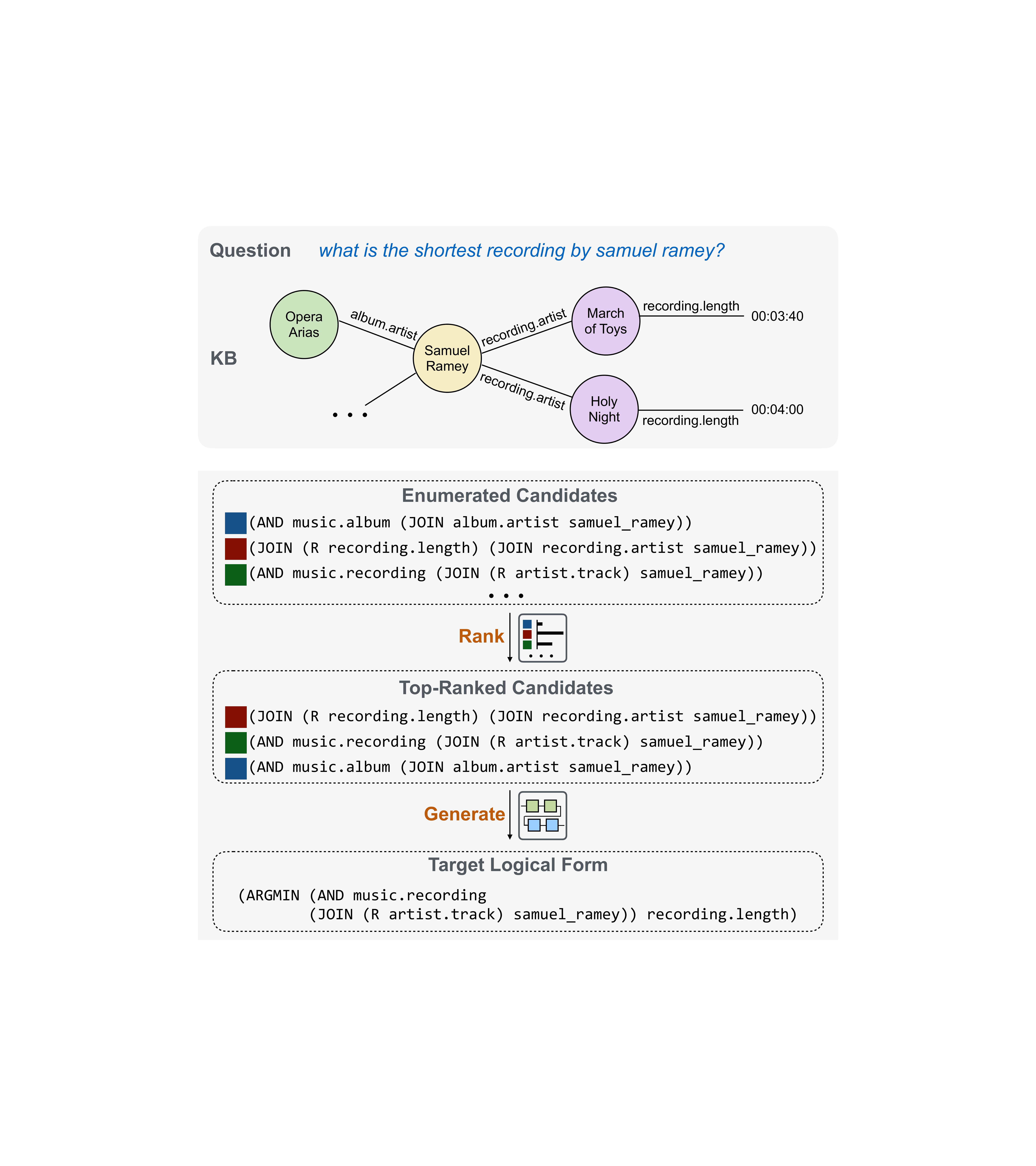}
    \caption{Overview of our rank-and-generate approach. Given a question, we first rank logical form candidates obtained by searching over the KB based on predefined rules. Here, the ground truth logical form is not in the top-ranked candidates as it is not covered by the rules. We solve this problem using another generation step that produces the correct logical form based on top-ranked candidates. The final logical form is executed over the KB to yield the answer.}
    \label{fig:framework}
\end{figure}

Modern knowledge bases (KB) are reliable sources of a huge amount of world knowledge but can be difficult to interact with since they are extremely large in scale and require specific query languages (e.g., Sparql) to access. Question Answering over Knowledge Base (KBQA) serves as a user-friendly way to query over KBs and has garnered increasing attention \cite{berant2013freebase,cai2013large}. Recent research has attempted to build systems achieving strong results on several public benchmarks that contain i.i.d. train and test distribution such as {\sc SimpleQ} \cite{simpleq} and {\sc WebQSP} \cite{webqsp}. However, users often want to ask questions involving unseen compositions or KB schema items (see Figure~\ref{fig:grail_exs} for examples), which still remains a challenge. Generation-based approaches (e.g., a seq-to-seq parser) are not effective enough to handle such practical generalization scenarios due to the difficulty of generating unseen KB schema items. Ranking-based approaches, which first generate a set of candidate logical forms using predefined rules and then select the best-scored one according to the question, have shown some success \cite{grail}. However, it suffers from the coverage problem, because it is often impractical to exhaust all the rules to cover the desired logical form due to the scale of the KB.

We propose \rngqa{}, a new framework targeted at generalization problems in the task of KBQA. Our approach combines a ranker with a generator, which addresses the coverage issue in ranking-only based approaches while still benefiting from their generalization power.
As shown in Figure~\ref{fig:framework}, we first employ a ranker to select a set of related logical forms from a pool of candidate logical forms obtained by searching over the graph.
The selected logical forms are not required to cover the correct one, but are semantically coherent and aligned with the underlying intents in the question. Next, we introduce a generator that consumes both the question and the top-k ranked candidates to compose the final logical form. The core idea of our approach is the interplay between the ranker and the generator: the ranker provides essential ingredients of KB schema items to the generator, which then further refines the top-candidates by complementing missing constructions or constraints, and hence allows covering a broader range of logical form space.

We base both our ranker and generator on pretrained language models for better generalization capability.
Unlike prior systems which rank candidates using a grammar-based parser \cite{berant2013freebase} or a seq-to-seq parser \cite{grail}, our ranker is a BERT-based \cite{bert} bi-encoder (taking as input question-candidate pair) trained to maximize the scores of ground truth logical forms while minimizing the scores of incorrect candidates. Such training schema allows learning from the contrast between the candidates in the entire territory, whereas prior parsing-based ranker \cite{berant2013freebase,grail} only learns to encourage the likelihood of the ground truth logical forms. We further develop an iterative-bootstrap-based training curriculum for efficiently training the ranker to distinguish spurious candidates (Section~\ref{sec:ranking}). 
In addition, we extend the proposed logical form ranker, keeping the architecture and logic the same, for the task of entity disambiguation, and show its effectiveness as a second-stage entity ranker.
Our generator is a T5-based \cite{2020t5} seq-to-seq model that fuses semantic and structural ingredients found in top-k candidates to compose the final logical form. To achieve this, we feed the generator with the question followed by a linearized sequence of the top-k candidates, which allows it to distill a refined logical form that will fully reflect the question intent by complementing the missing pieces or discarding the irrelevant parts without having to learn the low-level dynamics.

We test \rngqa{} on two datasets, \grail{} and \webqsp{}, and compare against an array of strong baselines. On \grail{}, a challenging dataset focused on generalization in KBQA, our approach sets the new state-of-the-art performance of 68.8 exact match 74.4 F1 score, surpassing prior SOTA (58.1 exact match and 65.3 F1 score) by a large margin. On the popular \webqsp{} dataset, \rngqa{} also outperforms the best prior approach (QGG \cite{qgg}) and achieves a new SOTA performance of 75.7 F1 score. The results demonstrate the effectiveness of our approach across all settings and especially in compositional generalization and zero-shot generalization.

    \section{Generation Augmented KBQA}

\subsection{Preliminaries}
A knowledge base collects knowledge data stored in the form of subject-relation-object triple ${(s,r,o)}$, where $s$ is an entity, $r$ is a binary relation, and $o$ can be entities or literals (e.g., date time, integer values, etc.).
Let the question be $x$, our task is to obtain a logical form $y$ that can be executed over the knowledge base to yield the final answer. 
Following \citet{grail}, we use \emph{s-expressions} to represent queries over knowledge base. S-expression (examples in Figure~\ref{fig:framework}) uses functions (e.g., \texttt{\small JOIN}) operating on set-based semantics and eliminates variable usages as in lambda DCS \cite{Liang2013LambdaDC}. This makes s-expression a suitable representation for the task of KBQA because it balances readability and compactness \cite{grail}.

\paragraph{Enumeration of Candidates}
Recall that our approach first uses a ranker model to score a list of candidate logical forms $C=\{c_i\}_{i=1}^m$ obtained via enumeration.
We'll first introduce how to enumerate the candidates before delving into the details of our ranking and generation models.

We start from every entity detected in the question and query the knowledge base for paths reachable within two hops. Next, we write down an s-expression corresponding to each of the paths, which constitutes a set of candidates. We note that we do not exhaust all the possible compositions when enumerating (e.g., we do not include comparative operations and argmin/max operations), and hence does not guarantee to cover the target s-expression. A more comprehensive enumeration method
is possible but will introduce a prohibitively large number (greater than 2,000,000 for some queries) of candidates. Therefore, it's impractical to cover every possible logical form when enumerating, and we seek to tackle this issue via our tailored generation model.

\subsection{Logical Form Ranking}
\begin{figure}[t]
    \centering
    \includegraphics[width=\linewidth,trim=330 320 330 290,clip]{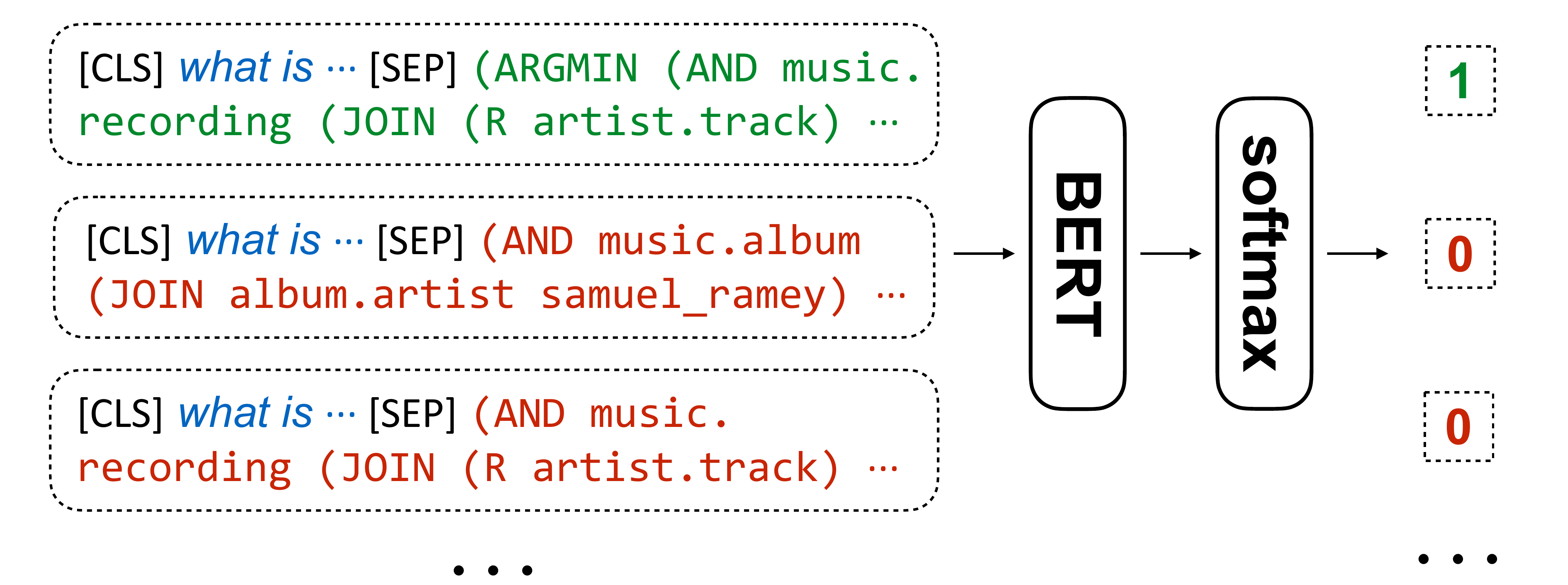}
    \caption{The ranker that learns from the contrast between the ground truth and negative candidates.}
    \label{fig:ranking_model}
\end{figure}

\label{sec:ranking}
Our ranker model learns to score each candidate logical form by maximizing the similarity between question and ground truth logical form while minimizing the similarities between the question and the negative logical forms (Figure~\ref{fig:ranking_model}). Specifically, given the question $x$ and a logical form candidate $c$, we use a BERT-based encoder that takes as input the concatenation of the question and the logical form and outputs a logit representing the similarity between them formulated as follows:

\begin{align*}
    s(x,y)=\textsc{Linear}(\textsc{BertCls}([x;y]))
\end{align*}

where $\textsc{BertCls}$ denotes the [CLS] representation of the concatenated input; $\textsc{Linear}$ is a projection layer reducing the representation to a scalar similarity score. The ranker is then optimized to minimize the following loss function:

\begin{equation}
\label{eq:obj_ranker}\mathcal{L}_{\text{ranker}}=-\frac{e^{s(x,y)}}{e^{s(x,y)} + \sum_{c\in C \land c\neq y}  e^{s(x,c)}}
\end{equation}

where the idea is to promote the ground truth logical form while penalizing the negative ones via a contrastive objective.
In contrast, the ranker employed in past work \cite{grail}, a seq-to-seq model, aims to directly map the question to target logical form, only leveraging supervision from the ground truth. Consequently, our ranker is more effective in distinguishing the correct logical forms from \emph{spurious ones} (similar but not equal to the ground truth ones).

\paragraph{Bootstrapping Negative Samples in Training}
Due to the large number of candidates and limited GPU memory, it is impractical to feed all the candidates $c\in C$ as in Eq~(\ref{eq:obj_ranker}) when training the ranker. Therefore, we need to sample a subset of negatives logical forms $C'\subset C$ at each batch. A naive way for sampling negative logical forms is to draw random samples. However, because the number of candidates is often large compared to the allowed size of negative samples in each batch, it may not be possible to cover spurious logical forms within the randomly selected samples.

We propose to sample negative logical forms by bootstrapping, inspired by the negative sampling methods used in \citet{densepr}. That is, we first train the ranker using random samples for several epochs to warm start it, and then choose the spurious logical forms that are confusing to the model as the negative samples for further training the model.
We find the ranker can benefit from this advanced negative sampling strategy and perform better compared to using random negative samples.

\subsection{Target Logical Form Generation}

\begin{figure}[t]
    \centering
    \includegraphics[width=\linewidth,trim=330 400 330 400,clip]{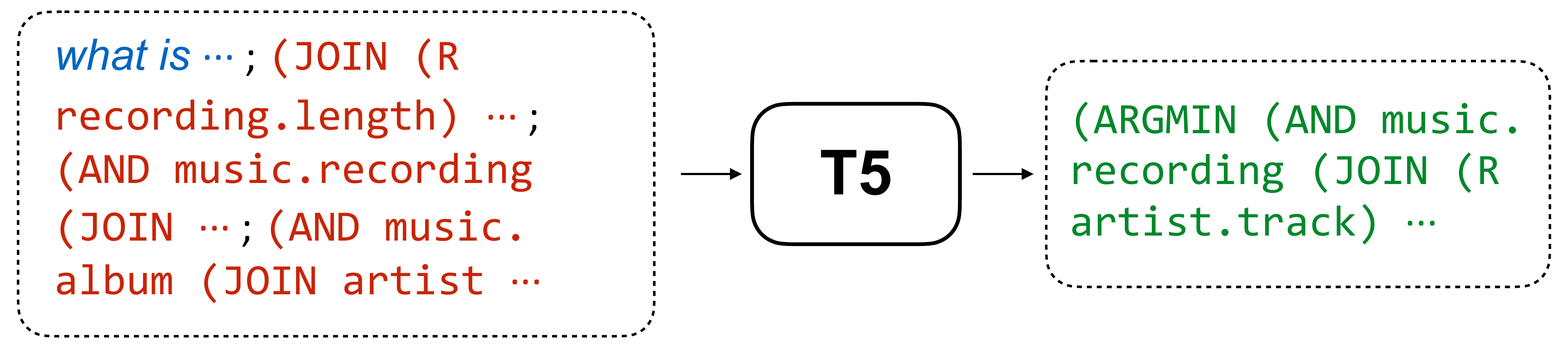}
    \caption{The generation model conditioned on question and top-ranked candidates returned by the ranker.}
    \label{fig:gen_model}
    \vspace{-0.5em}
\end{figure}

Having a ranked list of candidates, we introduce a generation model to compose the final logical form conditioned on the question and the top-k logical forms.
Our generator is a transformer-based seq-to-seq model \cite{vaswani2017transformer} instantiated from T5 (\cite{2020t5}), as it demonstrates strong performance in generation-related tasks.
As shown in Figure~\ref{fig:gen_model}, we construct the inputs by concatenating the question and the top-k candidates returned by the ranker separated by semi-colon (i.e., $[x ; c_{t_1} ; ... ; c_{t_k}]$). 
We train the model to generate the ground truth logical form autoregressively with cross-entropy objective using teacher forcing.
In the inference, we use beam-search to decode top-k target logical forms.
To construct the top-k logical form candidates needed for training the generator, we first train the ranker, and then use the rankings it produces on the training data.

Since the generation model can now leverage both the question and KB schema information (contained in the candidates), the context is much more specified as compared to only conditioning on the question. This enables our generator to leverage the training data more efficiently by focusing only on correcting or supplementing existing logical forms instead of learning both the generation rule and correctness of logical forms.

\begin{figure}[t]
    \centering
    \includegraphics[width=\linewidth,trim=350 240 350 240,clip]{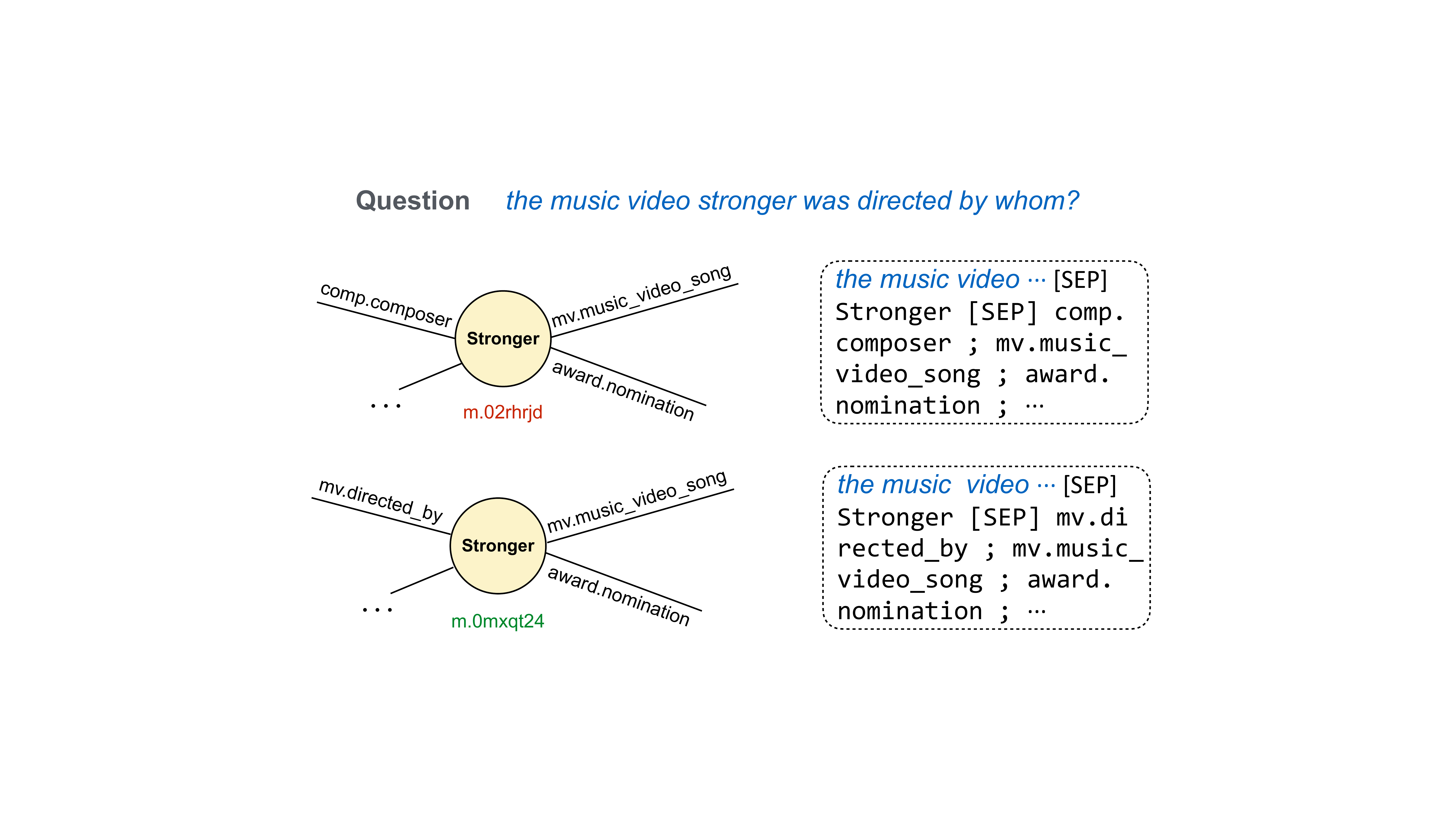}
    \caption{Illustrative example of running entity disambiguation as ranking. A confusing entity (red) and the correct entity (green) both match the surface form in the question. To distinguish them, we train an entity disambiguation model following the same architecture as in logical form ranking but construct inputs by concatenating the question and relations.}
    \label{fig:entity_disamb}
\end{figure}

\paragraph{Execution-Augmented Inference}
We use a vanilla T5 generation model without syntactic constraints, which does not guarantee the syntactic correctness nor executability of the produced logical forms. Therefore, we use an execution-augmented inference procedure, which is commonly used in prior semantic parsing related work \cite{robustfill,deepsketch}. We first decode top-k logical forms using beam search and then execute each logical form until we find one that yields a valid (non-empty) answer. In case that none of the top-k logical forms is valid, we return the top-ranked candidate obtained using the ranker as the final logical form, which is guaranteed to be executable. This inference schema can ensure finding one valid logical form for each problem. It is possible to incorporate a more complex mechanism to control the syntactic correctness in decoding (e.g., using grammar-based decoder \cite{asn} or dynamical beam pruning techniques \cite{structregex}).
We leave such extension aside since we find that executability of produced logical forms is not the bottleneck  (see Section~\ref{sec:analysis} in experiments).

\subsection{Extension: Entity Disambiguation as Ranking}
\label{sec:disamb}

Our ranking model is mainly proposed for the task of ranking candidate logical forms. Here, we introduce a simple way to adapt our ranking model for the task of entity disambiguation. A common paradigm of finding KB entities referred in a question is to first detect the entity mentions with an NER system and then run fuzzy matching based on the surface forms. This paradigm has been employed in various methods \cite{stagg,pullnet,retrack,grail}. One problem with this paradigm lies in entity disambiguation: a mention usually matches surface forms of more than one entities in the KB. 

A common way to disambiguate the matched entities is to choose the most popular one according to the popularity score provided by FACC1 project \cite{retrack,grail}, which can be imprecise in some cases. We show an example in Figure~\ref{fig:entity_disamb}. Consider the question ``the music video stronger was directed by whom?'' taken from \grail{}, where the most popular matched entity is ``Stronger'' ( \texttt{\small m.02rhrjd}, song by Kanye West)'' and the second is also ``Stronger'' (\texttt{\small m.0mxqqt24}, music video by Britney Spears). The surface form matching and popularity scores do not provide sufficient information needed for disambiguation.

However, it is possible to leverage the relation information linked with an entity to further help assess if it matches a mention in the question. By querying relations over KB,
we see there is a relation about mv director \texttt{\small mv.directed\_by} linking to 
\texttt{\small m.0mxqqt24}, but there are no such kind of relations connected with \texttt{\small m.02rhrjd}. We therefore cast the disambiguation problem to an entity ranking problem, and adapt the ranking model used before to tackle this problem. Given a mention, we concatenate the question with the relations for each entity candidate matching the mention.
We reuse the same model architecture and loss function as in Section~\ref{sec:ranking} to train another entity disambiguation model to further improve the ranking of the target entity. We apply our entity disambiguation model on \grail{}, and achieve substantial improvements in terms of entity linking.

    \begin{table*}[t]
    \centering
    \begin{tabular}{lcccccccc}
    \toprule
    & \multicolumn{2}{c}{Overall} & \multicolumn{2}{c}{I.I.D.} & \multicolumn{2}{c}{Compositional} & \multicolumn{2}{c}{Zero-Shot} \\\cmidrule{2-3} \cmidrule{4-5} \cmidrule{6-7} \cmidrule{8-9}
    & EM & F1 & EM & F1 & EM & F1 & EM & F1 \\
    \midrule
         QGG \cite{qgg} & $-$ & 36.7 & $-$ & 40.5 & $-$ & 33.0 & $-$ & 36.6 \\
    \midrule
        Bert Transduction \cite{grail} & 33.3 & 36.8& 51.8& 53.9& 31.0& 36.0& 25.7& 29.3 \\
         Bert Ranking \cite{grail} & 50.6 & 58.0 & 59.9 & 67.0 & 45.5 & 53.9 & 48.6 & 55.7 \\
         \midrule
         ArcaneQA (Anonymous) & 57.9 & 64.9 & 76.5 &  79.5 & 56.4 & 63.5 & 50.0 & 58.8 \\
         ReTrack \cite{retrack} & 58.1 & 65.3 &  84.4 &  87.5 & 61.5 & 70.9 & 44.6 & 52.5 \\
         S2QL (Anonymous) &  57.5 & 66.2 & 65.1 & 72.9 & 54.7	& 64.7 &55.1 &  63.6\\
         \midrule 
         RnG-KBQA (Ours) & \bf 68.8 & \bf 74.4 & \bf 86.2& \bf 89.0 & \bf 63.8 & \bf 71.2 & \bf 63.0 & \bf 69.2 \\
         \quad w/o Entity Disambiguation & 61.4 & 67.4 & 78.0 & 81.8& 55.0& 63.2& 56.7& 63.0\\
    \bottomrule
    \end{tabular}
    \caption{Exact match (EM) and F1 scores on the test split of \grail{}. The numbers of other approaches are taken from the leaderboard. \rngqa{} substantially outperforms prior methods by a large margin.}
    \label{tab:grail_main}
\end{table*}

\section{Experiments}
\label{sec:exp}
We mainly test our approach on \grail{} \cite{grail}, a challenging dataset focused on evaluating the generalization capabilities. We also experiment on \webqsp{} and compare against a number of prior approaches to demonstrate the general applicability of our approach.

\begin{figure}[t]
    \centering
    \includegraphics[width=\linewidth,trim=370 220 370 220,clip]{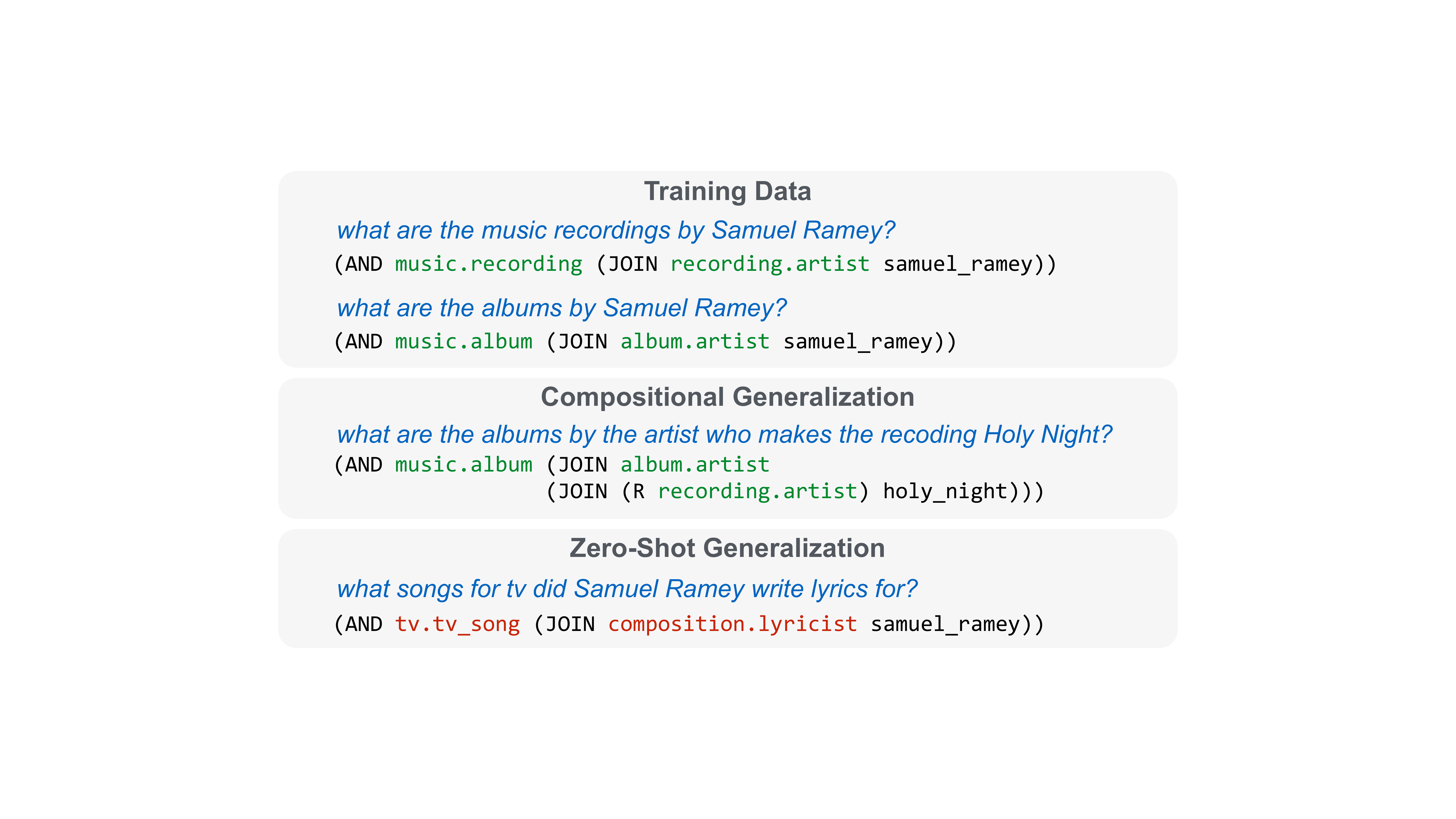}
    \caption{Examples of compositional generalization to new composition of KB schema items and zero-shot generalization to unseen schema items (red).}
    \label{fig:grail_exs}
\end{figure}

\subsection{Experiment: \grail{}}

\grail{} is the first dataset that evaluates the zero-shot generalization. Specifically, \grail{} contains 64,331 questions in total and carefully splits the data so as to evaluate three levels of generalization in the task of KBQA, including \emph{i.i.d.} setting, \emph{compositional} generalization to unseen composition, and  \emph{zero-shot} generalization to unseen KB schema (examples in Figure~\ref{fig:grail_exs}). The fraction of each setting in the test set is 25\%, 25\%, and 50\% , respectively. Aside from the generalization challenge, \grail{} also presents additional difficulty in terms of the large number of involved entities/relations, complex compositionality in the logical forms (up to 4 hops), and noisiness of the entities mentioned in questions \cite{grail}.

\paragraph{Implementation Detail}
We link an entity mention to an entity node in KB using our approach described in Section~\ref{sec:disamb}. We first use a \textsc{BERT-NER} systems provided by the authors of \grail{} to detect mention spans in the question. For each mention span, we match the span with surface forms in FACC1 project \cite{facc1}, rank the matched entities using popularity score, and retain the top-5 entity candidates. Lastly, we use the disambiguation model trained on \grail{} to select only one entity for each mention. Our entity ambulation model is initiated from BERT-base-uncased model provided by huggingface library \cite{huggingface}, and finetuned for 3 epochs with a learning rate of 1e-5 and a batch size of 8.

  When training the ranker, we sample 96 negative candidates using the strategy described in Section~\ref{sec:ranking}. Our ranker is finetuned from BERT-base-uncased for 3 epochs using a learning rate of 1e-5 and a batch size of 8. We do bootstrapping after every epoch. It is also noteworthy that we perform teacher-forcing when training the ranker, i.e., we use ground truth entity linking for training.

We base our generation model on T5-base \cite{2020t5}. We use top-5 candidates returned by the ranker and finetune for 10 epochs using a learning rate of 3e-5 and a batch size of 8.

\paragraph{Metrics} For \grail{}, we use exact match (EX) and F1 score (F1) as the metrics, all of which are computed using official evaluation script.

\paragraph{Results} Table~\ref{tab:grail_main} summarizes the results on \grail{}. The results of other approaches are directly taken from the leaderboard.\footnote{Accessed on 03/10/2022.} Overall, our approach sets the new state-of-the-art performance on \grail{} dataset, achieving 68.8 EM and 74.4 F1. This exhibits a large margin over the other approaches: our approach outperforms ReTrack \cite{retrack} by 10.7 EM and 8.2 F1.

Furthermore, \rngqa{} performs generally well for all three levels of generalization and is particularly strong in zero-shot setting. Our approach is slightly better than ReTrack and substantially better than all the other approaches in i.i.d. setting and compositional setting. However, ReTrack fails in generalizing to unseen KB Schema items and only achieves poor performance in zero-shot setting, whereas our approach is generalizable and beats ReTrack with a margin of 16.1 F1.

To directly compare the effectiveness of our rank-and-generate framework against rank-only baseline (BERT Ranking), we also provide the performance of a variant of \rngqa{} without the entity-disambiguation model. In this variant, we directly use the entity linking results provided by the authors of \citet{grail}. Under the same entity linking performance, our ranking-and-generation framework is able to improve the performance by 11.4 EM and 8.2 F1. Furthermore, the variant of our model without the entity-disambiguation module (RnG-KBQA w/o Entity Disambiguation) still substantially outperforms all other approaches. In particular, this variant beats ReTrack by 3.3 EM and 2.1 F1 even if ReTrack includes an entity disambiguation model that yields better entity linking performance. Please refer to Appendix~\ref{sec:app_a} for more discussion on entity linking performance.

\begin{table}[t]
    \centering
    \small
    \begin{tabular}{lccc}
    \toprule
    & F1 & EM & Hits\textsuperscript{@1} \\
    \midrule
    PullNet* \cite{pullnet} & 62.8 & $-$ & 67.8 \\
    GraftNet* \cite{graftnet} & $-$ & $-$ & 68.1 \\
    Bert Ranking* \cite{grail} & 67.0 & $-$ & $-$\\
    EmbedQA* \cite{saxena} &  $-$ & $-$ & 72.5 \\
    ReTrack* \cite{retrack} & 74.7 & $-$ & 74.6 \\
    \midrule
    Topic Units \cite{lan-2019-topic} & 67.9 & $-$ & 68.2  \\
    UHop \cite{chen-etal-2019-uhop} & 68.5 & $-$ & $-$ \\
    NSM \cite{liang-etal-2017-neural} & 69.0 & $-$ & $-$ \\
    ReTrack \cite{retrack} & 71.0 & $-$ & 71.6\\
    STAGG \cite{stagg}  &  71.7 & 63.9 & $-$ \\
    CBR \cite{casebase} & 72.8 & 70.0 & $-$ \\
    QGG \cite{qgg} & 74.0 & $-$ & $-$ \\
    \midrule
    \rngqa{} \textbf{(Ours)} & \bf 75.6 & \bf 71.1 & $-$\\
    \bottomrule
 \end{tabular}
    \caption{Results of \rngqa{} and baselines on \webqsp{}. * (approach in the top section) denotes using oracle entity linking annotations provided by the dataset. Our approach achieves the new state-of-the-art performance (75.6 F1) with a discernible margin over the performance of best prior method (74.0 F1 obtained by QGG). Our approach even outperforms a number of prior work using oracle entity linking annotations.}
    \label{tab:webqsp_main}
\end{table}

\subsection{Experiment: \webqsp{}}

\webqsp{} is a popular dataset which evaluates KBQA approaches in i.i.d. setting. It contains 4,937 question in total and requires reasoning chains with up to 2 hops. Since there is no official development split, we randomly sample 200 examples from the training set for validation.

\paragraph{Implementation Detail}
For experiments on \webqsp{}, we use ELQ \cite{li2020efficient} as the entity linker, which is trained on \webqsp{} dataset to perform entity detection and entity linking, since it produces more precise entity linking results and hence leads to less number of candidate logical forms for each question. Because ELQ always links a mention to only one entity, we do not need an entity-disambiguation step for \webqsp{} dataset. Similarly, we initiate the logical form ranker using BERT-base-uncased, and the generator using T5-base. We also sample 96 negative candidates for each question, and feed the top-5 candidates to the generation model. The ranker is trained for 10 epochs and we run bootstrapping every 2 epochs; the generator is trained for 20 epochs.

\paragraph{Metrics}
 F1 is used as the main evaluation metric. In addition, for approaches that are able to select entity sets as answers, we report the exact match (EM) used in the official evaluation. For information-retrieval based approaches that can only predict a single entity, we report Hits\textsuperscript{@1} (if the predicted entity is in the ground truth entity set), which is considered as a loose approximation of EM.
 
 \paragraph{Results} For baseline approaches, we directly take the results reported in corresponding original paper. As shown in Table~\ref{tab:grail_main}, \rngqa{} achieves 75.6 F1, surpassing the prior state-of-the-art (QGG) by 1.6. Our approach also achieves the best EM score of 71.1, surpassing CBR \cite{casebase}. The performance of our approach obtained using ELQ-predicted entity linking outperforms all the prior methods, even if they are allowed to use oracle entity linking annotations (denoted as * in the top section). It is also noteworthy that both CBR and QGG, the two methods achieving strong performance closest to ours, use an entity linker with equal or better performance compared to ours. In particular, CBR also uses ELQ for entity linking. QGG uses an entity linker achieving 85.2 entity linking F1 (calculated using public available code) which is slightly better than ours achieving 84.8 entity linking F1. To summarize, the results on \webqsp{} suggest that, in addition to outstanding generalization capability, our approach is also as strong in solving simpler questions in i.i.d. setting.

\subsection{Analysis}
\label{sec:analysis}

\begin{table}[t]
    \centering
    \small
    \begin{tabular}{lcc}
    \toprule
    & \grail{} & \webqsp{} \\
    \midrule
    Full Model & 75.1 & 75.6 \\
    \midrule
    Gen Only (Rand Rank) & 47.6 & 69.9  \\
    \midrule
    Rank Only & 69.8 & 72.7 \\
    Rank Only (w/o Bootstrap) & 68.6 & 71.3 \\
    \bottomrule
    \end{tabular}
    \caption{F1 scores on \grail{} (dev set) and \webqsp{} of three ablations, including a generation-only variant (Gen Only, which uses randomly selected logical form candidates), a ranking-only variant (Rank Only), and a ranking-only variant without using bootstrap training strategy (w/o Bootstrap). Removing either component leads to performance deterioration.}
    \label{tab:ablation}
\end{table}

\paragraph{Ablation Study}
We first compare the performance of our full model against incomplete ablations in Table~\ref{tab:ablation}. We derive a generation-only (Gen Only) model from our base model by replacing the trained ranker with a random ranker, which leads to a performance drop of 27.5 and 5.7 on \grail{} and \webqsp{}, respectively. The performance deterioration is especially sharp on \grail{} as it requires generalizing to unseen KB schema items, for which the  generator typically needs to be based on a good set of candidates to be effective.

To test the effects of our generation step, we compare the performance of a ranking-only variant (directly using the top-ranked candidate) against the performance of the full model. As shown in Table~\ref{tab:ablation}, the generation model is able to remedy some cases not addressable by the ranking model alone, which boosts the performance by 5.3 on \grail{} and 2.9 on \webqsp{}.

We additionally evaluate the performance of a ranking model trained without bootstrapping strategy introduced in Section~\ref{sec:ranking}. The performance of this variant lags its counterpart by 1.2 and 1.4 on \grail{} and \webqsp{}, respectively. The bootstrapping strategy is indeed helpful for training the ranker to better distinguish spurious candidates.

\begin{figure}[t]
    \centering
    \includegraphics[width=\linewidth,trim=260 60 260 75,clip]{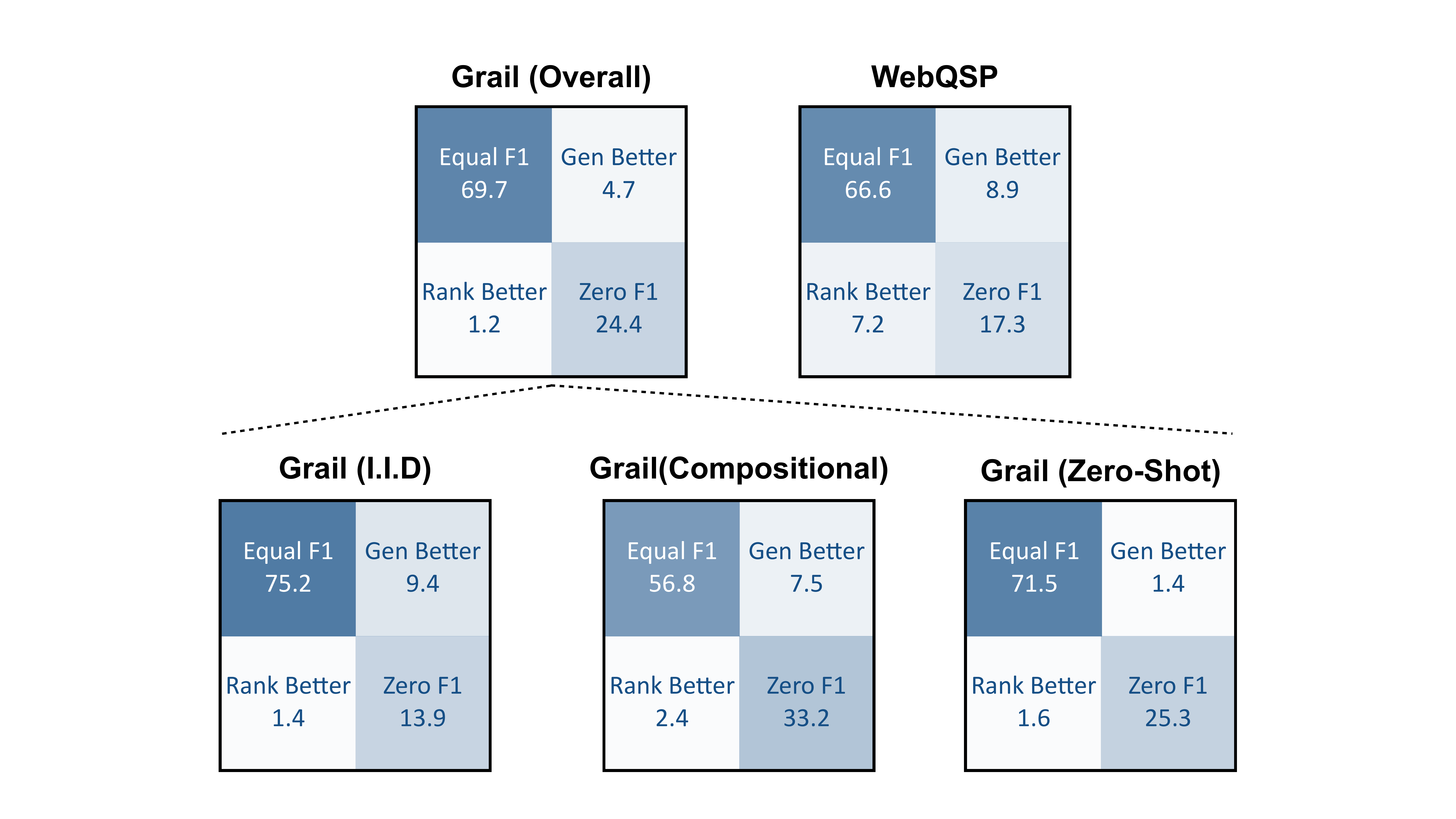}
        \caption{Comparison between the ranker's top predictions and the generator's top predictions. Generation model mostly keeps or improves the prediction while occasionally introducing errors.}
    \label{fig:comp_matrices}
\end{figure}

\begin{figure}[t]
    \centering
    \small
    \renewcommand{\tabcolsep}{1mm}
    \begin{tabularx}{\linewidth}{|r X|}
    \hline
    \multicolumn{2}{|c|}{\bf Generation Better Than Ranking} \\
    \hline
    (a) Q & \textit{what is the shortest recording by samuel ramey?}\\
     R & \texttt{\small (AND music.recording (JOIN recording.artist ramey))}  \\
    G & \texttt{\small (ARGMIN (AND music.recording (JOIN recording.artist ramey)) recording.length)}\\
    \hline
    (b) Q & \textit{where did kevin love go to college?}\\
     R & \texttt{\small (JOIN education.institution (JOIN person.education love))}\\
     G & \texttt{\small (AND (JOIN topic.notable\_types college) (JOIN edu.institution (JOIN  person.education love)))}  \\
    \hline
        \multicolumn{2}{|c|}{\bf Ranking Better Than Generation} \\
    \hline
    (c) Q & what song for tv or television did benny davis compose? \\
     R &  \texttt{\small (AND tv.tv\_song (JOIN composition.lyricist davis)) }  \\
    G & \texttt{\small (AND tv.tv\_song (JOIN composition.song (JOIN composition.composer  davis))) }  \\
     \hline
     (d) Q & what team does heskey play for? \\
     R &  \texttt{\small (JOIN sports\_team\_roster.team (JOIN pro\_athlete.teams heskey)) }  \\
    G & \texttt{\small (JOIN sports\_team\_roster.team (AND (JOIN sports\_team\_roster.from 2015) (JOIN pro\_athlete.teams heskey))) }  \\
    \hline
    \end{tabularx}
    \caption{Examples of outputs from the generator (G) and ranker (R). A generation step is able to compensate some missing operators not supported in the enumeration (a), or patch some implicit clue (b). However, generator does introduce errors if it produces another prediction when there is inherent ambiguity in the question and the top-ranked one is indeed correct (c). Generator also adds unnecessary constraint sometimes (d).}
    \label{fig:gen_exs}
    \vspace{-1em}
\end{figure}

\paragraph{Comparing Outputs of Ranking Model and Generation Model}

We have demonstrated the benefit of adding a generation stage on top of the ranking step on previous result sections. Here, we present a more detailed comparison between the outputs of ranking model and generation model. 
Figure~\ref{fig:comp_matrices} presents the ``comparison matrices'' showing the fractions of questions where

$\circ$ top left: the top ranking prediction and top generation prediction achieves a equal nonzero F1,

$\circ$ top right: the top generation prediction is better,

$\circ$ bottom left: the top ranking prediction is better,

$\circ$ bottom right: they both fail (achieving a 0 F1).

The generator retains the ranking predictions without any modifications for most of the time. For 4.7\% and 8.9\% of the questions from \grail{} and \webqsp{}, respectively, the generator is able to fix the top-ranked candidates and improves the performance. 
Although generator can make mistakes in non-negligible fraction of examples on \webqsp{}, it is mostly caused by introducing false constraints (e.g., Figure~\ref{fig:gen_exs} (d)). Thanks to our execution-guided inference procedure, we can still turn back to ranker-predicted results when the generator makes mistakes, which allows tolerating generation errors to some extent.

We also show the break down by types of generalization on \grail{} (bottom row in Figure~\ref{fig:comp_matrices}). Generation stage is more helpful in i.i.d. and compositional setting, but less effective in zero-shot setting, as it involves unseen relations that are usually hard to generate.

\begin{table}[t]
    \centering
    \begin{tabular}{l c c c c}
    \toprule
   \multirow{2}{*}{} & \multicolumn{2}{c}{\grail{}} & \multicolumn{2}{c}{\webqsp{}} \\
       & EXEC & VALID & EXEC & VALID \\
     \midrule
      Top-1 & 99.7 & 88.1 &  98.7 & 91.1 \\
      Top-3 & 99.7 & 89.4 &  99.5 & 94.5 \\
      Top-5 & 99.7 & 89.8 &  99.5 & 94.6 \\
     Top-10 & 99.7	& 90.4& 99.5 & 95.4 \\
      \bottomrule
    \end{tabular}
    \caption{The chances of finding an executable (EXEC) and a valid (VALID) logical form among the top-k generated. logical forms.}
    \label{tab:exec_rate}
\end{table}

\paragraph{Executability}
We use executability to further measure the quality of generated outputs. Table~\ref{tab:exec_rate} shows executable rate (producing an executable logical forms) and valid rate (producing a logical form that yields non-empty answer) among the top-k decoded list. Nearly all the top-1 logical forms are executable. This suggests that the generation model can indeed produce high-quality predictions in terms of syntactic correctness and consistency with KB. As the beam size increases, more valid logical forms can be found in the top-k list, which our inference procedure can benefit from.

\paragraph{Output Examples of Ranking Model and Generation Model} For more intuitive understanding of how the generator works, we attach several concrete examples (Figure~\ref{fig:gen_exs}). As suggested by example (a), the generation model can remedy some missing operations (\textit{\small ARGMIN}) not supported when enumerating. It can also patch the top-ranked candidate with implicit constraints: the \texttt{\small (JOIN topic.notable\_types college)} in (b) is not explicitly stated, and our NER system fails to recognize \emph{college} as an entity.

As in example (c), the generation model makes a worse prediction sometimes because it prefers another prediction in the top-ranked list due to inherent ambiguity in the question. It can also fail when falsely adding a constraint which results in empty answer (d).

    \section{Related Work}

KBQA is a promising technique for users to efficiently query over large KB, which has been extensively studied over the last decade. Past work has collected a series of datasets \cite{webqsp, simpleq,metaq,graphq, grail} as well as proposed a diversity of approaches for this task.

One line of KBQA approaches first constructs a query-specific subgraph with information retrieved from the KB and then rank entity nodes to select top entities as the answer \cite{graftnet,pullnet,saxena,refinedqa,transfernet}. The subgraph can either be retrieved in one-shot using heuristic rules \cite{graftnet}, or iteratively built using learned models \cite{pullnet, transfernet, refinedqa, saxena}. Later, a neural model operating over subgraph is employed to determine the answer nodes \cite{graftnet, pullnet, transfernet}. Such information retrieval based approaches are usually less interpretable as they do not produce the inference path reaching the answer, whereas our approach is more transparent since we are able to produce logical forms.

More closely related to our approach, another line answers a question by parsing it into an executable logical form in various representations, including lambda-DCS \cite{Liang2013LambdaDC,berant2013freebase}, sparql query \cite{casebase}, graph query \cite{stagg,graphq,qgg}, and s-expression \cite{grail}. Past work has attempted to generate logical forms using grammar-based parsera \cite{berant2013freebase} or seq-to-seq parsers \cite{zhang2019-complex}. There has also been an alternative way that first enumerates a list of logical form candidates and then choose one that best matches the intents in the question \cite{qgg,luo-etal-2018-knowledge,stagg,yavuz-etal-2016-improving,yavuz-etal-2017-recovering,reddy-etal-2017-universal,SunZ0Q20}. Our approach differs in that we employ a generation stage to remedy the coverage issue which these approaches often suffer from.

    \section{Conclusion}
We have presented \rngqa{} for question answering over knowledge base. \rngqa{} consists of a ranking step and a generation step. Our ranker trained with iterative bootstrapping strategy can better distinguish correct logical forms from spurious ones than prior seq-to-seq ranker. Our generator can further remedy uncovered operations or implicitly mentioned constraints in the top-ranked logical forms. The experimental results on two datasets, \grail{} and \webqsp{}, suggest the strong performance of our approach: \rngqa{} achieves new state-of-the-art performance on both datasets, and particularly outperforms prior methods in generalization setting by a large margin.

\section*{Acknowledgments}

Thanks to Greg Durrett and Yasumasa Onoe for their valuable suggestions. Thanks to Man Luo, Haopeng Zhang, and everyone at Salesforce Research for helpful discussions, as well as to the anonymous
reviewers for their helpful feedback. 

    
    \bibliography{custom}
    \bibliographystyle{acl_natbib}
    
    \onecolumn
\newpage
\twocolumn

\appendix

\section{Details of Entity Linking Performance}
\label{sec:app_a}
         
\begin{table}[t]
    \centering
    \small
    \begin{tabular}{l c c}
    \toprule
         &  Linking F1 & KBQA F1\\
    \midrule
      Bert Ranking \cite{grail}   &  72.2 & 58.0 \\
       ReTrack \cite{retrack}  & 77.4 & 65.3 \\
    \midrule
        RnG-KBQA (Ours)  &  79.6 & 74.4 \\
      \quad w/o Entity Disambiguation   &  72.2 & 67.4 \\
     \bottomrule
    \end{tabular}
    \caption{The entity linking F1 (on dev set) and the corresponding final F1 scores (on leaderboard) on \grail{} of various methods.}
    \label{tab:el_results}
\end{table}

Table~\ref{tab:el_results} shows the entity linking performance and KBQA performance on \grail{} of various methods. Compared to the popularity-based baseline (Bert Ranking), Our entity disambiguation model is effective and successfully improves the entity linking F1 by 7.4, which boosts the final KBQA F1 score by 7.0. Our entity linking model is also better than the Bootleg approach \cite{Orr2021BootlegCT} used in ReTrack \cite{retrack}.

Furthermore, our method without the entity disambiguation modules outperforms Bert Ranking with a substantially large margin (11.4 F1 score). Our method even beats ReTrack when it is built upon a much better entity linking model. The results suggest the strong effectiveness of our rank-and-generate framework.

\end{document}